\PassOptionsToPackage{table}{xcolor}
\documentclass[10pt,twocolumn,letterpaper]{article}

\usepackage[pagenumbers]{cvpr} % To force page numbers, e.g. for an arXiv version

\usepackage{multirow}
\usepackage{svg}
\usepackage{dsfont}
\usepackage{graphicx} % For \resizebox
\usepackage{booktabs}
\definecolor{cvprblue}{rgb}{0.21,0.49,0.74}
\usepackage[pagebackref,breaklinks,colorlinks,allcolors=cvprblue]{hyperref}

\def\paperID{5} % *** Enter the Paper ID here
\def\confName{CVPR}
\def\confYear{2025}

\title{On the Robustness of GUI Grounding Models Against Image Attacks}

\author{Haoren Zhao\\
Hangzhou Dianzi University\\
{\tt\small zhaohaoren2020@outlook.com}
\and
Tianyi Chen\\
Microsoft\\
{\tt\small tiachen@microsoft.com}
\and
Zhen Wang\\
Hangzhou Dianzi University\\
{\tt\small wangzhen@hdu.edu.cn}
}

\begin{document}
\maketitle
\begin{abstract}
Graphical User Interface (GUI) grounding models are crucial for enabling intelligent agents to understand and interact with complex visual interfaces. However, these models face significant robustness challenges in real-world scenarios due to natural noise and adversarial perturbations, and their robustness remains underexplored. In this study, we systematically evaluate the robustness of state-of-the-art GUI grounding models, such as UGround, under three conditions: natural noise, untargeted adversarial attacks, and targeted adversarial attacks. Our experiments, which were conducted across a wide range of GUI environments, including mobile, desktop, and web interfaces, have clearly demonstrated that GUI grounding models exhibit a high degree of sensitivity to adversarial perturbations and low-resolution conditions. These findings provide valuable insights into the vulnerabilities of GUI grounding models and establish a strong benchmark for future research aimed at enhancing their robustness in practical applications. Our code is available at \url{https://github.com/ZZZhr-1/Robust_GUI_Grounding}.
\end{abstract}    
\section{Introduction}
\label{sec:intro}

% Graphical User Interface (GUI) agents are designed to automate operations based on user instructions, enhancing human-computer interaction efficiency and improving the overall user experience \cite{wu2024copilot}. Recently, multimodal large language models (MLLMs) have achieved remarkable progress in visual grounding capabilities, opening new avenues for the development of GUI agent systems \cite{wang2024qwen2,achiam2023gpt}. By fine-tuning MLLMs on GUI grounding tasks, these models have demonstrated impressive performance in accurately locating target elements within complex interfaces using visual information and natural language instructions. \cite{cheng2024seeclick,wu2025osatlas,gou2025uground}.
Graphical User Interface (GUI) agents are designed to automate operations based on user instructions, enhancing human-computer interaction efficiency and improving the overall user experience~\cite{wu2024copilot}. Recently, multimodal large language models (MLLMs) have achieved remarkable progress in visual grounding capabilities, opening new avenues for the development of GUI agent systems \cite{wang2024qwen2,achiam2023gpt}. By fine-tuning MLLMs on GUI grounding tasks, these models have demonstrated impressive performance in accurately locating target elements within complex GUI using visual information and natural language instructions~\cite{cheng2024seeclick,wu2025osatlas,gou2025uground}.

\begin{figure*}
    \centering
    \includegraphics[width=1\linewidth]{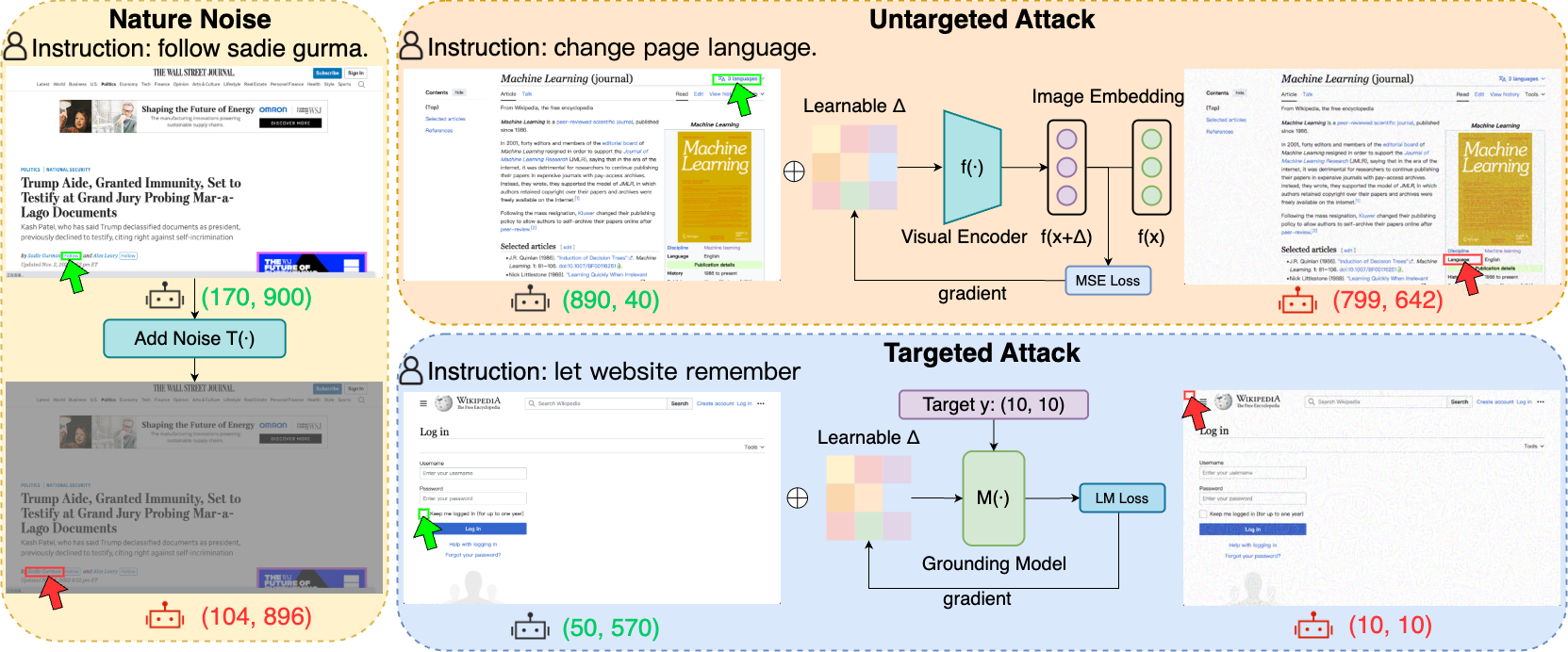}
    \caption{Examples of natural noise (color jitter), untargeted attack, and targeted attack (results on the Uground-V1 model).}
    \label{fig:overview}
\end{figure*}

Despite their potential, GUI grounding models face significant robustness challenges in open and real-world settings \cite{wu2024agentattack}. Sensitivity to input variations can result in incorrect responses under malicious or abnormal conditions, posing risks to system stability and security \cite{chen2025aeia,chiang2025web}. Visual inconsistencies caused by differences in devices, such as varying operating systems or screen resolutions, can further lead to grounding errors. Moreover, adversaries can exploit crafted perturbations to mislead models, potentially directing agents to malicious links or websites \cite{zhang2024attacking}.

Although there has been some progress in recent years on the robustness of multimodal models, most research has focused on tasks like visual question answering (VQA) and image captioning~\cite{zhao2023evaluating,fu2023misusing,dong2023robust}, with relatively little research on visual grounding tasks~\cite{gao2024adversarial}. Moreover, GUI grounding has unique scene characteristics, such as non-natural images, diverse interface element layouts, the complexity of icon types, and small object detection. These unique features present additional challenges for ensuring GUI grounding robustness. Therefore, a deeper investigation into the robustness of GUI grounding models in complex environments is crucial for improving their stability and security in real-world applications.

% In this work, we investigate the robustness of the latest GUI grounding models across various attack scenarios, focusing on three key aspects: 1) Robustness under natural noise (e.g., resolution changes and image blurring); 2) Untargeted attacks on the image encoder, where adversarial perturbations disrupt feature outputs, leading to incorrect grounding results; 3) White-box targeted attacks, where perturbations direct the model to click a designated 0.04\% target region, smaller than most icons and text, ensuring the attack’s significance. Our extensive evaluations of GUI grounding models in complex environments (e.g., mobile, desktop, web) offer valuable insights for future research and practical applications. An attack sample is shown in Figure 1.
In this work, we investigate the robustness of the latest GUI grounding models across various attack scenarios, focusing on three key aspects: \textit{(i)} robustness under natural noise (\eg, resolution changes and image blurring); \textit{(ii)} untargeted attacks on the image encoder, where adversarial perturbations disrupt feature outputs, leading to incorrect grounding results; and \textit{(iii)} white-box targeted attacks, where perturbations direct the model to click a designated 0.04\% target region, smaller than most icons and text, ensuring the attack’s significance. Our extensive evaluations of GUI grounding models in varying environments (\eg, mobile, desktop, web) offer valuable insights for future research and practical applications. Figure~\ref{fig:overview} illustrates our attack method and some examples of the attack results.

Our main contributions can be summarized as follows:
\begin{itemize}[leftmargin=*,noitemsep, topsep=0pt]
\item We systematically analyze the robustness of GUI grounding models under various perturbation conditions.
\item We experimentally validate the performance of GUI grounding models in scenarios that involve natural noise, untargeted attacks, and targeted attacks.
\item We establish an essential and reliable experimental benchmark to advance future research and applications in GUI grounding robustness.
\end{itemize}
%-------------------------------------------------------------------------

\section{Related work}
\label{sec:formatting}

%-------------------------------------------------------------------------
\subsection{GUI Agents and GUI Grounding Models}

In the field of GUI agents, large language models (LLMs) and multimodal large language models (MLLMs) have demonstrated significant potential \cite{kim2023language, wang2024mobile, xie2024osworld}. Many multimodal agents rely on HTML or a11y trees for grounding \cite{koh2024visualwebarena, gur2024real, zheng2024gpt}, while lacking generality. In contrast, some studies have explored pixel-level, visually grounded GUI agents \cite{shaw2023pixels, zhang2024you, hong2024cogagent}. Due to the significant differences between GUI and natural scenes, traditional visual grounding methods often perform poorly in GUI contexts \cite{cheng2024seeclick}. The Set-of-Mark (SoM) method \cite{yang2023set} introduces visual markers (\eg, boxes and numbers) to guide models in identifying target objects. However, it heavily depends on complete object information or segmentation \cite{zheng2024gpt, koh2024visualwebarena, ma2024groma}. The SeeClick model \cite{cheng2024seeclick} fine-tuned Qwen-VL \cite{bai2023qwen} on GUI data, establishing a new grounding benchmark. SeeAct \cite{gou2025uground} proposed a two-stage approach that separates planning from visual grounding, achieving strong performance in benchmark tests. OS-Atlas \cite{wu2025osatlas} developed a multi-platform data collection framework and designed a dedicated large-action model for GUI agents. Despite these advancements, concerns regarding the security of deploying large language model agents in real-world applications remain an open problem~\cite{ngo2022alignment, ruan2023identifying, mo2024trembling, wudissecting}.

%-------------------------------------------------------------------------
\subsection{Robustness for MLLMS}

Machine learning models are vulnerable to adversarial examples, and small perturbations to the input can lead to incorrect predictions \cite{goodfellow2014explaining,biggio2013evasion,szegedy2013intriguing,dong2018boosting,zhang2019theoretically,wang2022transferable}. A large number of studies have been carried out to improve adversarial attacks and defenses~\cite{carlini2017towards,madry2018towards,raghunathan2018certified,cohen2019certified}. Early research mainly focused on image classifiers, and later studies extended adversarial attacks to large language models~\cite{jia2017adversarial,wallace2019universal} and adversarial attacks on multi-modal large language models~\cite{dong2023robust,gao2024inducing,wang2024stop}. Recent research has explored the adversarial robustness in application scenarios such as visual question answering (VQA)~\cite{fu2023misusing} and image caption~\cite{zhao2023evaluating,cui2024robustness}. However, the adversarial robustness of multi-modal large language models with visual grounding capabilities has not been fully explored~\cite{gao2024adversarial}, especially in the GUI field. To this end, we have designed a variety of attack methods to evaluate the robustness of GUI Grounding models. 
\section{Methodology}

\subsection{Preliminary}

The GUI grounding model predicts an element's location \(y\) (bounding box or coordinate point) from a screenshot \(s\) and description \(x\). The numerical digits are directly processed as tokens, and the MLLMs are trained with standard autogressive loss. The Grounding can be considered as success if the predicted position \(y\) falls within the ground-truth bounding box of the corresponding element. 

\noindent
\textbf{Threat Models.} For natural noise, the model faces threats from various factors such as different operating systems, themes, resolutions, and renderers, which introduce boundary blur, color variations, etc. These noises are injected based on predefined distributions rather than being adversarially constructed. Specifically, we evaluate the sensitivities of models to different noise types to induce incorrect outputs. For both untargeted and targeted attacks, an adversary aims to optimize an imperceptible perturbation to construct an adversarial image $x'$ for attack purposes. Following previous works~\cite{qi2023visual,dong2023robust,gao2024adversarial}, we assume that the adversary has full or partial access to the victim model and restricts the perturbation within a predefined $l_{\infty}$ norm bound $\epsilon$ to ensure it remains undetectable.

\subsection{Robustness Under Natural Noise}

% In the GUI Grounding task, the robustness of a model under natural noise can be evaluated by introducing real-world disturbances. Given an interface screenshot $s$ and an element description $x$, we apply a noise transformation $T(\cdot)$ to $s$, resulting in the transformed input $s' = T(s)$. Let the GUI Grounding model be defined as $M(s, x) \to \hat{y}$, where $\hat{y}$ represents the predicted element position (either normalized coordinates or bounding boxes). The robustness of the model can be measured using the Grounding Success Rate (SR) metric:

% \begin{equation} SR = \mathbb{E}_{(s, x, y) \sim \mathcal{D}} \big[1[\hat{y}' \in B(y)] \big], \end{equation}
% where $\hat{y}'$ is the predicted position for the transformed input $s'$, i.e., $\hat{y}' = M(s', x)$, $B(y)$ denotes the ground truth bounding box of the element, and $1$ is an indicator function. 

In the GUI Grounding task, the robustness of a model under natural noise can be evaluated by introducing real-world disturbances. Given an interface screenshot $s$ and an element description $x$, we apply a noise transformation $T(\cdot)$ to $s$, resulting in the transformed input $s' = T(s)$. Let the GUI Grounding model be defined as $\mathcal{M}(s, x) \to \hat{y}$, where $\hat{y}$ represents the predicted element position (either normalized coordinates or bounding boxes). A prediction is deemed correctly grounded if the predicted position lies within the confines of the ground-truth bounding box. Then model robustness is measured by the grounding success rate (SR),
\begin{equation} SR = \mathbb{E}_{(s, x, y) \sim \mathcal{D}} \big[\mathds{1}[\hat{y}' \in B(y)] \big], \end{equation}
where $\hat{y}'$ is the predicted position for the transformed input $s'$, \ie, $\hat{y}' = M(s', x)$, $B(y)$ denotes the ground truth bounding box of the element, and $\mathds{1}$ is an indicator function. 

\subsection{Untargeted Adversarial Attacks}

Multimodal language models (MLLMs) typically use a visual encoder $f(\cdot)$ to extract image embeddings, which are then combined with text embeddings and fed into a large language model (LLM). When the attacker has access to the model's visual encoder, untargeted attacks can be carried out by maximizing the $l_2$ distance between the image embeddings of the original image $s$ to produce the adversarial image $\hat{s} = s + \delta$. In particular, the adversarial sample is constructed by optimizing the following objective function,

\begin{equation}
\max_{\delta} \|f(s + \delta) - f(s)\|_2^2 \quad \text{subject to} \quad \|\delta\|_\infty \leq \epsilon,
\end{equation}
where $\delta$ is the adversarial perturbation, and the constraint $\|\delta\|_\infty \leq \epsilon$ ensures that the perturbation does not exceed $\epsilon$ in pixel-wise changes. The image embedding of the adversarial sample diverges from that of the clean sample, causing the model to fail in making correct predictions.

\subsection{Targeted Adversarial Attacks}

In targeted attacks, the attacker aims to construct an adversarial perturbation $\delta$ such that the GUI Grounding model $M(s + \delta, x)$ outputs the target location $t$. We assume the attacker has full access to the model, so the attack can be achieved by minimizing the language model (LM) loss between the model's output and the target text. The optimization objective function is formulated as,
\begin{equation}
\max_{\delta} \sum_{k=1}^{K} \log P(t_k \mid t_{<k}, s+\delta, x; \theta) \quad \text{subject to} \quad \|\delta\|_\infty \leq \epsilon,
\end{equation}
where $P(t_k \mid t_{<k}, s + \delta, x; \theta)$ is the model's probability of generating the target token $t_k$ at the $k$-th step, and $\theta$ represents the model parameters and $x$ denotes the instruction. The constraint $\|\delta\|_\infty \leq \epsilon$ ensures that the perturbation remains visually imperceptible.
\section{Experiments}

\subsection{Experimental Setups}

% \textbf{Models and Datasets.} We consider the latest GUI Grounding models as targets for attack: SeeClick~\cite{cheng2024seeclick}, OS-Atlas-Base-7B~\cite{wu2025osatlas}, and UGround-V1-7B (Qwen2-VL-based)~\cite{gou2025uground}. All models are of 7B scale. Additionally, we use the ScreenSpot-V2~\cite{wu2025osatlas,cheng2024seeclick} dataset for evaluation, which includes samples from Mobile, Desktop, and Web environments.
\textbf{Models and Datasets.} We consider the latest GUI Grounding models as targets for attack: SeeClick~\cite{cheng2024seeclick}, OS-Atlas-Base-7B~\cite{wu2025osatlas}, and UGround-V1-7B (Qwen2-VL-based)~\cite{gou2025uground}. All models are selected as about 7B scale to balance inference cost and quality. Additionally, we use the ScreenSpot-V2~\cite{wu2025osatlas,cheng2024seeclick} dataset for evaluation, which includes samples from Mobile, Desktop, and Web environments, encompassing both textual and icon targets.

% since the benchmark GUI elements are  unaffected by sensor noise, motion blur, or perspective distortions~\cite{zhao2022ood,mao2023coco,wang2023survey}, 

\noindent
% \textbf{Baselines and Setups.} First, to simulate real-world UI perturbations such as sensor noise, motion blur, and perspective distortion, we introduce scaling, blurring, color jitter, and contrast shifts. Specifically, we evaluate input resolutions ranging from 1024$\times$768 to 256$\times$768, apply Gaussian blur with a kernel size of 9, add Gaussian noise with mean 0 and standard deviation 30, reduce contrast by 0.5, and apply color jitter with an amplitude of 0.5.
\textbf{Baselines and Setups.} First, to simulate real-world UI perturbations and evaluate the model’s robustness, we exclude noise like lighting changes, perspective transformations, and random rotations, as GUIs are not affected by sensors or distortions. Specifically, we introduce Gaussian noise, Gaussian blur, color jitter, and contrast adjustments, and evaluate inputs with different maximum pixel value constraints to better assess the model’s adaptability.

Second, for adversarial attacks, we adopt the 100-step PGD algorithm~\cite{madry2018towards}. Following prior work~\cite{qi2023visual,dong2023robust,gao2024adversarial}, we use an $l_{\infty}$ constraint with a perturbation budget of $\epsilon = 16$ and a step size of $\alpha = 1$. In untargeted attacks, only the visual encoder is accessible, whereas targeted attacks assume full model access. The target area is the top-left 0.04\% of the image without loss of generality. Since the OS-Atlas-Base-7B model outputs bounding boxes rather than precise coordinates, its target $y$ is defined as a bounding box. We evaluate the models under both high-resolution and low-resolution conditions.

% \textbf{Baselines and Setups.} We consider three different types of attacks. At first, we introduce scaling, blurring, color jitter, and contrast shifts to reflect real-world UI perturbations of sensor noise, motion blur, and perspective distortions. We evaluate inputs from 1024x768 to 256x768 and apply commonly used Gaussian blur with a kernel size of 9, Gaussian noise with a mean of 0 and standard deviation of 30, contrast reduction by 0.5, and color jitter with an amplitude of 0.5.

% Secondly, for adversarial attacks, we employ a 100-step PGD algorithm~\cite{madry2018towards}. Following the setup in previous works~\cite{qi2023visual,dong2023robust,gao2024adversarial}, we use an $l_{\infty}$ constraint, setting the perturbation $\epsilon = 16$ and the optimization step size $\alpha = 1$. In untargeted attacks, the visual encoder is accessible, whereas in targeted attacks, full access is granted. The top-left 0.04\% region is selected as the target area. Since the OS-Atlas-Base-7B model cannot effectively generate coordinates and instead produces bounding boxes, its target
% $y$ is set as a bounding box, while for other models, the target $y$ is defined as a coordinate point. We evaluate the models under both high-resolution and low-resolution conditions.

\noindent
\textbf{Evaluation Metrics.} To evaluate the models' robustness to natural noise, we use the Success Rate (SR) as the evaluation metric. A prediction is considered successful if the predicted center of the coordinates or bounding box falls within the ground truth bounding box. A higher SR indicates better robustness. For untargeted attacks, we use the Attack Success Rate (ASR), which measures the proportion of the decrease in the model's SR after being attacked. For targeted attacks, ASR is defined as the success rate of predictions falling within the target area. A higher ASR indicates a more effective attack.

\subsection{Main Results}

\begin{table*}[h!]
\centering
\caption{Attack success rates of untargeted attacks on ScreenSpot-v2 for the three models at high and low resolutions. }
\label{tab:untarget_attack}
\resizebox{0.9\textwidth}{!}{%
\begin{tabular}{lllcccccccc}
\toprule
\multirow{2}{*}{\textbf{Resolution}} & \multirow{2}{*}{\textbf{Model}} & 
\multirow{2}{*}{\textbf{Setting}} & \multicolumn{2}{c}{\textbf{Mobile}} & \multicolumn{2}{c}{\textbf{Desktop}} & \multicolumn{2}{c}{\textbf{Web}} & \textbf{Avg} \\
\cmidrule(r){4-5} \cmidrule(r){6-7} \cmidrule(r){8-9}
& & & \textbf{Text} & \textbf{Icon/Widget} & \textbf{Text} & \textbf{Icon/Widget} & \textbf{Text} & \textbf{Icon/Widget} &  \\
\midrule
\multirow{4}{*}{High} 
& \multirow{2}{*}{OS-Atlas-Base-7B} & \cellcolor{gray!20} No Attack &  \cellcolor{gray!20} 94.14 & \cellcolor{gray!20} 72.99 & \cellcolor{gray!20} 92.78 & \cellcolor{gray!20} 66.43 & \cellcolor{gray!20} 89.32 & \cellcolor{gray!20} 78.32 & \cellcolor{gray!20} 82.33 \\
& & Untargeted & 36.27 & 59.75 & 62.77 & \textbf{74.20} & 53.11 & 67.93 & 59.00 \\
\cmidrule(r){2-10}
\multirow{12}{*}{Low} 
& \multirow{2}{*}{Uground-V1} & \cellcolor{gray!20} No Attack & \cellcolor{gray!20} 96.21 & \cellcolor{gray!20} 83.89 & \cellcolor{gray!20} 94.85 & \cellcolor{gray!20} 75.71 & \cellcolor{gray!20} 91.88 & \cellcolor{gray!20} 78.33 & \cellcolor{gray!20} 86.81 \\
& & Untargeted & 22.58 & 44.07 & 45.66 & \textbf{66.04} & 29.30 & 40.26 & 41.32 \\
\midrule
& \multirow{2}{*}{SeeClick} & \cellcolor{gray!20} No Attack & \cellcolor{gray!20} 78.62 & \cellcolor{gray!20} 48.82 & \cellcolor{gray!20} 73.20 & \cellcolor{gray!20} 29.29 & \cellcolor{gray!20} 59.83 & \cellcolor{gray!20} 22.66 & \cellcolor{gray!20} 52.07 \\
& & Untargeted & 64.03 & 76.71 & \textbf{86.63} & 73.16 & 79.29 & 56.53 & 72.73 \\
\cmidrule(r){2-10}
& \multirow{2}{*}{OS-Atlas-Base-7B} & \cellcolor{gray!20} No Attack & \cellcolor{gray!20} 72.41 & \cellcolor{gray!20} 41.23 & \cellcolor{gray!20} 48.45 & \cellcolor{gray!20} 26.43 & \cellcolor{gray!20} 43.16 & \cellcolor{gray!20} 37.44 & \cellcolor{gray!20} 44.85 \\
& & Untargeted & 39.52 & 60.93 & 67.02 & 62.16 & \textbf{82.18} & 68.43 & 63.37 \\
\cmidrule(r){2-10}
& \multirow{2}{*}{Uground-V1} & \cellcolor{gray!20} No Attack & \cellcolor{gray!20}94.48 & \cellcolor{gray!20}70.14 & \cellcolor{gray!20}85.05 & \cellcolor{gray!20}55.00 & \cellcolor{gray!20}65.81 & \cellcolor{gray!20}50.74 & \cellcolor{gray!20}70.20 \\
& & Untargeted & 55.84 & 58.78 & 73.33 & 83.11 & \textbf{84.41} & 69.91 & 70.90 \\

\bottomrule
\end{tabular}%
}
\end{table*}

\begin{table*}[h!]
\centering
\caption{Attack success rates of targeted attacks on ScreenSpot-v2 for the three models at high and low resolutions.}
\label{tab:target_attack}
\resizebox{0.9\textwidth}{!}{%
\begin{tabular}{llcccccccc}
\toprule
\multirow{2}{*}{\textbf{Resolution}} & \multirow{2}{*}{\textbf{Model}} & \multicolumn{2}{c}{\textbf{Mobile}} & \multicolumn{2}{c}{\textbf{Desktop}} & \multicolumn{2}{c}{\textbf{Web}} & \textbf{Avg} \\
\cmidrule(r){3-4} \cmidrule(r){5-6} \cmidrule(r){7-8}
& & \textbf{Text} & \textbf{Icon/Widget} & \textbf{Text} & \textbf{Icon/Widget} & \textbf{Text} & \textbf{Icon/Widget} &  \\
\midrule
\multirow{2}{*}{High} 
& OS-Atlas-Base-7B & 0.00 & 0.47 & 0.00 & 0.71 & 0.43 & \textbf{0.98} & 0.43 \\
& Uground-V1 & 2.07 & 7.58 & 13.92 & \textbf{25.71} & 3.42 & 6.40 & 9.85 \\
\midrule
\multirow{3}{*}{Low} 
& SecClick & 35.52 & 53.55 & 90.20 & \textbf{95.71} & 91.88 & 91.13 & 76.33 \\
& OS-Atlas-Base-7B & 2.07 & 4.27 & 2.06 & \textbf{7.14} & 2.99 & 3.45 & 3.66 \\
& Uground-V1 & 5.52 & 15.17 & 30.41 & \textbf{38.57} & 20.09 & 30.05 & 23.30 \\
\bottomrule
\end{tabular}%
}
\end{table*}

\begin{figure}
  \centering
  \begin{subfigure}{0.49\linewidth}
    \includegraphics[width=1\linewidth]{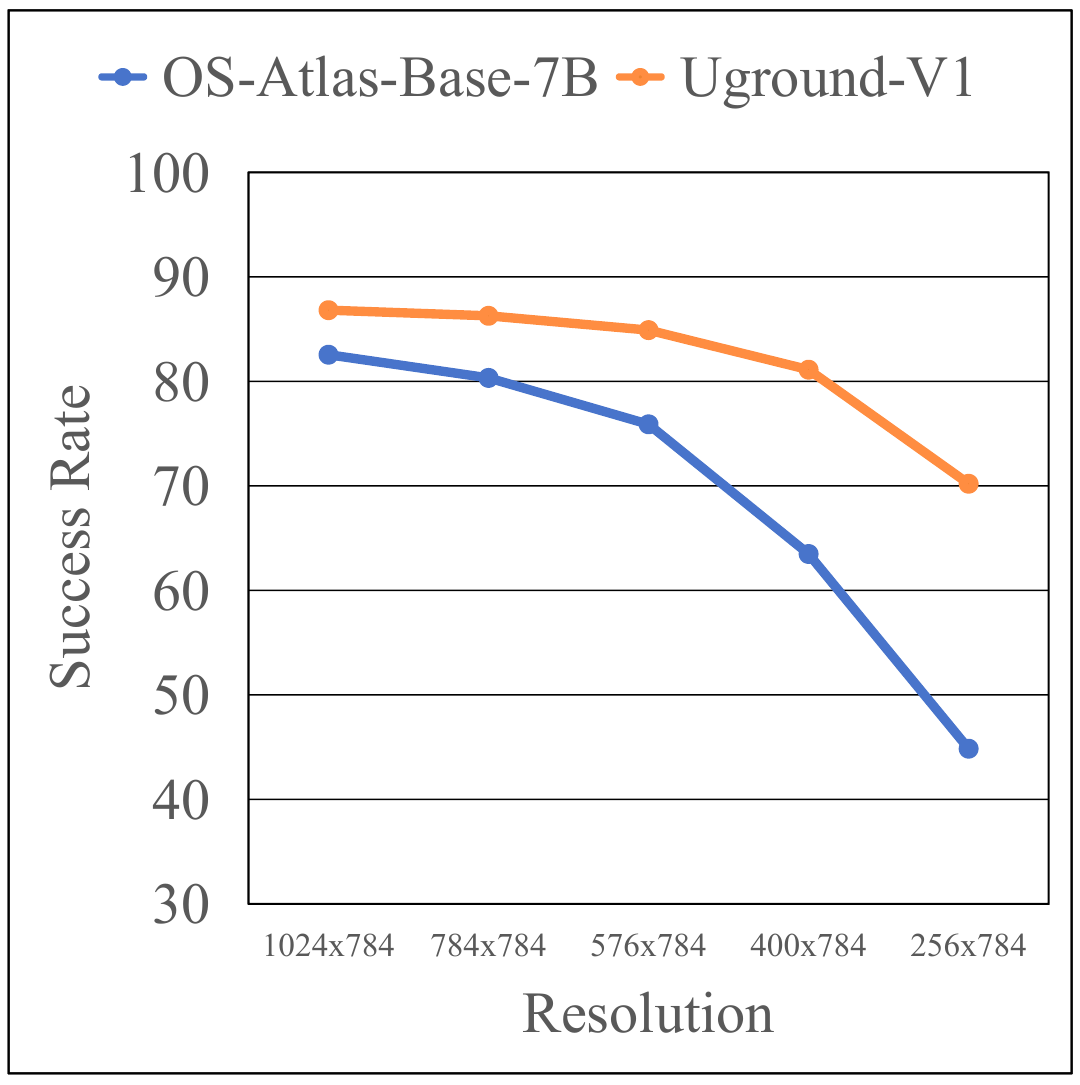}    
    \caption{Performance across resolutions.} 
    \label{fig:res}
  \end{subfigure}
  \hfill
  \begin{subfigure}{0.49\linewidth}
    \includegraphics[width=1\linewidth]{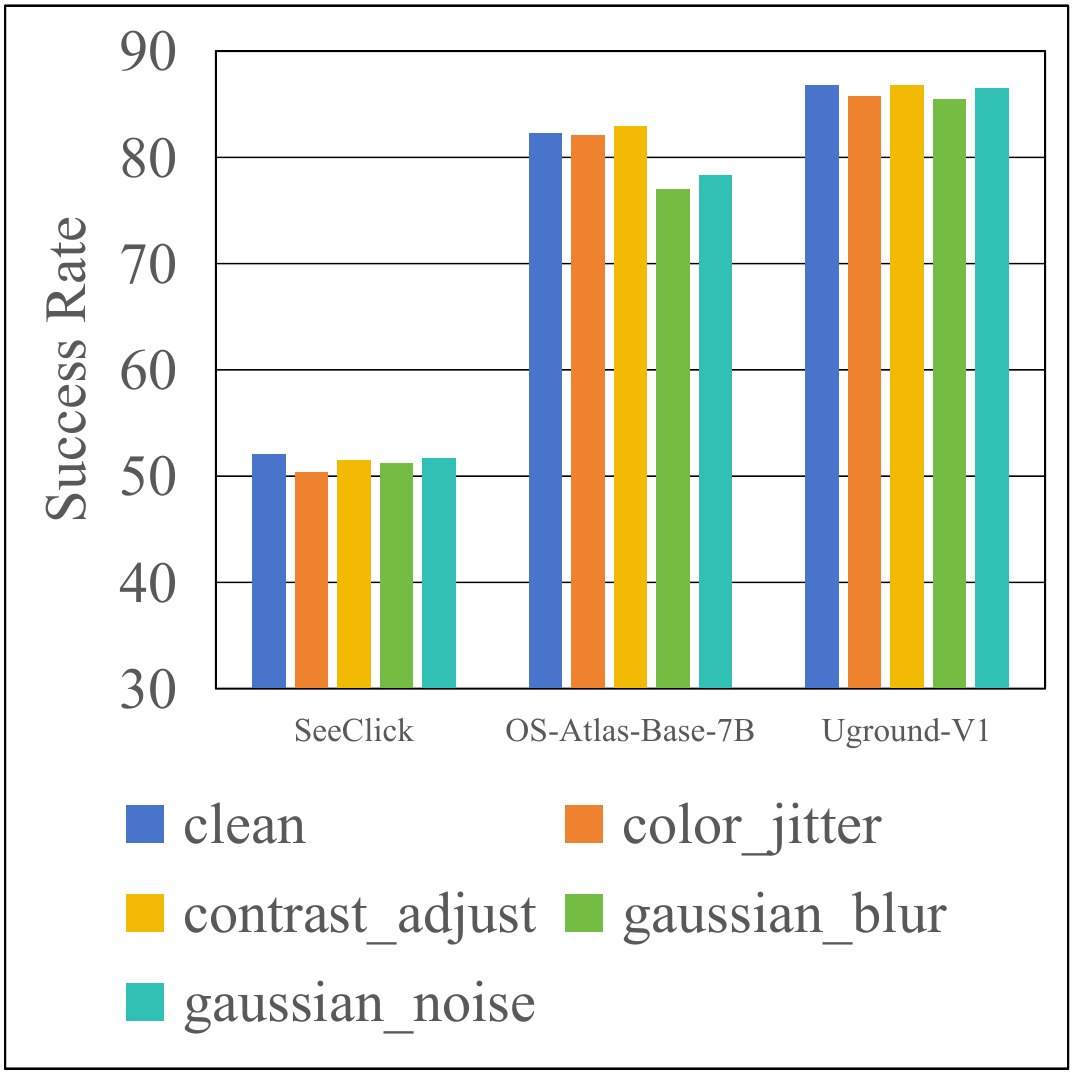}    
    \caption{Performance with natural noise.} 
    \label{fig:noise}
  \end{subfigure}
  \caption{Average performance of models under different resolutions and natural noise.}
  \label{fig:noise_res}
\end{figure}

Figure~\ref{fig:res} shows the average performance of OS-Atlas-Base-7B and UGround-V1 under varying pixel values. Both models degrade as maximum pixel value decreases, with OS-Atlas-Base-7B and UGround-V1 scoring 44.85\% and 70.20\% at 256x784 pixels (equivalent to SeeClick) respectively. UGround-V1 demonstrates better robustness to low resolution. Figure \ref{fig:noise} presents model performance under four types of natural noise, where OS-Atlas-Base-7B experiences the largest drop (5.27\%) under Gaussian blur.

% \begin{figure}
%     \centering
%     \includegraphics[width=1\linewidth]{pictures/noise_res.pdf}    
%     \caption{Average performance of models under different resolutions and natural noise.}
%     \label{fig:attack_result}
% \end{figure}

% Table 1 summarizes the untargeted attack success rates of the three models in different scenarios under both high and low resolutions. The results show that, compared to high-resolution conditions, the attack success rates increase significantly under low-resolution conditions. The highest attack success rate for each model is highlighted in bold. The models demonstrate the best robustness in the Mobile scenario, likely due to the simpler interface of mobile environments.
Table~\ref{tab:untarget_attack} summarizes the untargeted attack success rates of the three models in different scenarios under both high and low resolutions. The results show that, compared to high-resolution conditions, the attack success rates increase significantly under low-resolution conditions. The highest attack success rate for each model is highlighted in bold. Models exhibit the greatest robustness in the Mobile scenario, likely due to the simpler interfaces and streamlined design characteristic of mobile environments.

Table~\ref{tab:target_attack} reports the targeted attack success rates of the three models across different scenarios under high and low resolutions. Under low resolution, the attack success rate of SeeClick based on Qwen-VL is significantly higher than that of the other models based on Qwen2-VL. The OS-Atlas-Base-7B model exhibits the lowest targeted attack success rate at 3.66\%, possibly because attacks targeting bounding boxes (bbox) are more challenging. Under high-resolution conditions, the targeted attack success rates for the OS-Atlas-Base-7B and UGround-V1 models are relatively low. Notably, the Icon task in the desktop environment achieves the highest attack success rate.

\section{Conclusion}
% This paper investigates the robustness of GUI grounding models under natural noise, untargeted, and targeted adversarial attacks. Through extensive experiments across mobile, desktop, and web environments, we find that while these models show some resilience to natural noise, they are notably vulnerable to low-resolution inputs and adversarial perturbations. We hope our findings provide a benchmark for robustness evaluation and inspire future research on developing stronger visual grounding techniques for multi-modal large language models.
In this paper, we investigate the robustness of GUI grounding models under natural noise, untargeted, and targeted adversarial attacks. Through extensive experiments across mobile, desktop, and web environments, we find that while these models show some resilience to natural noise, they are notably vulnerable to low-resolution inputs and carefully crafted adversarial perturbations. We hope our findings can serve as a benchmark for evaluating the robustness of GUI grounding models and inspire future research toward developing more reliable and robust GUI grounding~\mbox{techniques}.
{
    \small
    \bibliographystyle{ieeenat_fullname}
    \bibliography{main}
}


\end{document}